\documentclass[sigconf]{acmart}
\usepackage{booktabs, multirow, tabularx}
\AtBeginDocument{%
  }

\setcopyright{acmlicensed}
\copyrightyear{2025}
\acmYear{2025}
\acmDOI{XXXXXXX.XXXXXXX}

\acmConference[Conference acronym 'XX]{Make sure to enter the correct
  conference title from your rights confirmation email}{June 03--05,
  2025}{City, State}
\acmBooktitle{Proceedings of the ...}
\acmISBN{978-1-4503-XXXX-X/25/06}

\begin{document}

\title{Automated Clinical Problem Detection from SOAP Notes using a Collaborative Multi-Agent LLM Architecture}

\author{Yeawon Lee}
\email{yl3427@drexel.edu}
\orcid{0009-0009-4209-2672}
\affiliation{%
  \institution{Drexel University}
  \city{Philadelphia}
  \state{PA}
  \country{USA}
}

\author{Xiaoyang Wang}
\email{xw388@drexel.edu}
\orcid{0000-0002-8471-4670}

\affiliation{%
  \institution{Drexel University}
  \city{Philadelphia}
  \state{PA}
  \country{USA}
}

\author{Christopher C. Yang}
\email{chris.yang@drexel.edu}
\orcid{0000-0001-5463-6926}
\affiliation{%
  \institution{Drexel University}
  \city{Philadelphia}
  \state{PA}
  \country{USA}
}

\renewcommand{\shortauthors}{Lee et al.}

\begin{abstract}
Accurate interpretation of clinical narratives is critical for patient care, but the complexity of these notes makes automation challenging. While Large Language Models (LLMs) show promise, single-model approaches can lack the robustness required for high-stakes clinical tasks. We introduce a collaborative multi-agent system (MAS) that models a clinical consultation team to address this gap. The system is tasked with identifying clinical problems by analyzing only the Subjective (S) and Objective (O) sections of SOAP notes, simulating the diagnostic reasoning process of synthesizing raw data into an assessment. A Manager agent orchestrates a dynamically assigned team of specialist agents who engage in a hierarchical, iterative debate to reach a consensus. We evaluated our MAS against a single-agent baseline on a curated dataset of 420 MIMIC-III notes. The dynamic multi-agent configuration demonstrated consistently improved performance in identifying congestive heart failure, acute kidney injury, and sepsis. Qualitative analysis of the agent debates reveals that this structure effectively surfaces and weighs conflicting evidence, though it can occasionally be susceptible to groupthink. By modeling a clinical team's reasoning process, our system offers a promising path toward more accurate, robust, and interpretable clinical decision support tools.
\end{abstract}



\maketitle

\section{Introduction}
The integration of Large Language Models (LLMs) into healthcare is accelerating, offering powerful tools for analyzing the vast amounts of text generated in clinical care. These clinical narratives are often unstructured and complex, yet they contain critical information reflecting a clinician's reasoning process. LLM-based agents show promise in automatically interpreting this data, raising hopes for their use in real clinical settings.

While prompt-based, single-agent LLMs have shown utility in clinical Natural Language Processing (NLP), they can suffer from a single, unverified pathway of reasoning. This creates a single point of failure. Concurrently, LLM-powered multi-agent systems (MAS) have emerged as a new paradigm, demonstrating sophisticated capabilities in complex problem-solving. By enabling mechanisms like debate and role-playing, MAS can enhance robustness and mitigate the risk of reasoning failures, as the task can be readily reassigned to other agents \cite{tran2025multiagentcollaborationmechanismssurvey}.

We apply this multi-agent paradigm to the clinical domain, proposing a system designed to function as a digital consultation team. Our work addresses a critical, yet under-explored task: identifying specific clinical problems from the Subjective and Objective (S+O) sections of SOAP notes in the MIMIC-III dataset. If we included the Assessment (A) or Plan (P) section, the task could become a simple keyword search, as the diagnosis is often explicitly stated there. By limiting the input to the S+O data, we challenge the system to synthesize raw information and reason implicitly, a task central to the clinical thought process.

Our framework's novelty lies in its specific, hierarchical architecture designed to mimic a real-world clinical consultation. The system features a manager agent that dynamically assembles a team of specialists tailored to the clinical problem at hand. These specialists then engage in an iterative debate to reach a consensus. By combining insights from multiple 'perspectives' (agents) and refining conclusions through a debate mechanism, our system improves its reasoning and predictive accuracy through collaborative error correction, aligning with general findings that collective agent outputs can surpass individual model capabilities \cite{tran2025multiagentcollaborationmechanismssurvey, wang2024mixtureofagentsenhanceslargelanguage, talebirad2023multiagentcollaborationharnessingpower}.

Furthermore, the framework offers enhanced explainability. Instead of a single "black-box" answer, it is possible to trace the reasoning, disagreements, and consensus-building process among agents. Our qualitative analysis of these debate transcripts provides concrete examples of how this collaborative process corrects errors and, in some cases, how it can lead to flawed group dynamics. 

Our primary research question is: Does a collaborative multi-agent architecture with dynamically assigned specialists and a hierarchical debate protocol outperform a single-LLM baseline in identifying clinical problems from S+O notes? 

The main contributions of this work are:
\begin{itemize}
    \item A novel MAS architecture for clinical problem identification, featuring dynamic specialist assignment and a hierarchical, iterative debate mechanism.
    \item An empirical demonstration of the system's effectiveness, supported by a qualitative analysis that reveals the mechanisms of both successful collaboration and its failure modes.
    \item A detailed analysis of the emergent agent collaboration dynamics, including specialist recruitment patterns and the identification of influential agent roles.
    \item A publicly available pipeline that utilizes a locally-hosted LLM, demonstrating a privacy-preserving approach for handling sensitive clinical data while ensuring reproducibility on the MIMIC-III dataset.
\end{itemize}

This paper is structured as follows: We first review related work in LLMs in healthcare and multi-agent systems. We then detail our methodology, including dataset construction and system architecture. Next, we present our results, followed by a discussion of their implications and limitations. Finally, we conclude and suggest directions for future work.

\section{Related Work}
\subsection{LLMs in Healthcare}
Large language models (LLMs) have emerged as transformative tools in healthcare, leveraging their advanced natural language processing (NLP) capabilities to address a wide range of applications, from clinical decision support to patient education and biomedical research. Singhal et al.~\citep{singhal2023large} demonstrated that GPT-4 encodes clinical knowledge effectively, achieving performance comparable to human experts on medical question-answering benchmarks like MedQA, with an accuracy of over 80\% on complex clinical queries. Similarly, Wu et al.~\citep{wu2024pmc} introduced PMC-LLaMA, a fine-tuned LLaMA model on medical papers, which outperformed general-purpose LLMs in tasks like named entity recognition and relation extraction from clinical texts. Multimodal LLMs (M-LLMs), which integrate text with other data modalities such as medical images and time-series data, have also gained traction. A comprehensive review by AlSaad et al.~\citep{alsaad2024multimodal} explored M-LLMs' potential in real-time patient monitoring and predictive analytics, such as detecting early signs of sepsis from vital sign data in intensive care units.

\subsection{LLM-Powered Multi-Agent Systems}
LLMs can serve as the core reasoning engine or "brain" for autonomous agents capable of inference and decision-making, showong emergent behaviors \cite{tran2025multiagentcollaborationmechanismssurvey}. These LLM-based agents can collaborate to solve complex problems. A comprehensive survey by Tran et al. \cite{tran2025multiagentcollaborationmechanismssurvey} characterizes these systems by their collaboration structures (e.g., centralized, decentralized), agent roles, and interaction protocols. Our work utilizes a centrally-coordinated structure with a dedicated Manager agent. General MAS frameworks like Mixture-of-Agents (MoA) have shown that structured interaction among multiple LLMs can lead to state-of-the-art performance on benchmark tasks by leveraging diverse perspectives for iterative refinement \cite{wang2024mixtureofagentsenhanceslargelanguage}. Talebirad and Nadiri \cite{talebirad2023multiagentcollaborationharnessingpower} also propose a general framework where agents with different roles and capabilities interact to achieve a common goal.

\subsection{MAS in the Medical Domain}
The application of MAS in medicine is an emerging field. A notable example is Agent Hospital \cite{li2025agenthospitalsimulacrumhospital}, a large-scale simulation where numerous agents representing doctors, nurses, and patients interact to model the entire patient treatment lifecycle. This work demonstrates the potential of MAS for complex medical process modeling. While systems like Agent Hospital focus on broad, longitudinal simulations, the application of MAS to the specific and highly detailed task of interpreting individual clinical documents like SOAP notes is less explored. Our work fills this gap by designing and evaluating a MAS tailored for collaborative reasoning on a single, information-dense clinical note to solve a specific diagnostic question, customizing general MAS concepts for the challenges of clinical text analysis.

\subsection{The Structure and Significance of SOAP Notes in Clinical NLP}

The SOAP note format, first developed by Dr. Lawrence Weed as part of the Problem-Oriented Medical Record (POMR), is a widely adopted standard for clinical documentation \cite{Weed1968}. It organizes clinical information into four distinct sections: Subjective (the patient's self-reported history and symptoms), Objective (measurable data such as vital signs, physical exam findings, and lab results), Assessment (the clinician's diagnosis or analysis), and Plan (the proposed treatment).

More than just a documentation template, the SOAP note's structure is significant because it reflects the fundamental process of clinical reasoning. A clinician gathers subjective and objective data to form a diagnostic assessment, upon which a plan is built. This progression from raw data (S+O) to synthesized conclusion (A) represents the core of diagnostic inference. We identified this as a critical and less-explored opportunity for automated systems.

Much of the prior work in clinical NLP that aims to identify patient problems has relied on information extraction from the full note \cite{WANG201834}, often by simply finding diagnoses explicitly stated in the Assessment section \cite{gao-etal-2022-hierarchical}. While useful, this approach sidesteps the more complex challenge of reasoning. Our work deliberately focuses on the S+O to 'A' transition. By providing only the Subjective and Objective sections as input, we force our system to perform a task analogous to genuine clinical reasoning: synthesizing disparate pieces of information to arrive at a diagnostic conclusion. This makes the task a more authentic test of an LLM's analytical capabilities and is central to our research contribution.

\section{Methodology}
Our methodology centers on a multi-agent system specifically designed to perform clinical problem identification from the Subjective and Objective (S+O) sections of SOAP notes. We first describe the process used to construct our specialized dataset from the corpus created by Gao et al. \cite{gao-etal-2022-hierarchical}, followed by the detailed architecture of our system and the experimental setup used for evaluation.

\subsection{Dataset Construction for Clinical Reasoning}
Our study began with a corpus of 768 MIMIC-III progress notes from Gao et al. \cite{gao-etal-2022-hierarchical}, which contains line-by-line annotations classifying text into the SOAP format. While this annotated corpus provides a valuable foundation, we implemented a multi-step filtering process to create a more focused dataset that could  evaluate the diagnostic reasoning capabilities of our multi-agent system.

\paragraph{Step 1: Curation of a Clinically Relevant Problem Set}
The original problem list was broad, including not only complex diagnoses but also simple symptoms. To focus our task on conditions requiring diagnostic inference, we curated a master list of 14 substantial diagnoses (e.g., 'myocardial infarction', 'sepsis', 'acute kidney injury').

\paragraph{Step 2: Selecting for Implicit Diagnoses}
The primary goal of our system is to simulate clinical reasoning, not to perform simple keyword searches. To achieve this, we filtered the notes based on a critical criterion: a note was included for a given problem only if the diagnosis was present in the ground-truth summary but explicitly absent from the text of its Subjective (S) and Objective (O) sections. This step was designed to eliminate "easy" cases where the diagnosis is plainly stated in the input text.

\paragraph{Step 3: Final Dataset Assembly}
We aggregated all unique notes that met this filtering criterion for at least one of the 14 curated problems. This process yielded our final dataset of 420 unique progress notes, each representing a challenging case that requires genuine clinical inference.

\paragraph{Task Formulation}
For the experiments in this paper, we focused on the three most frequent and clinically significant problems within our final dataset: "congestive heart failure," "sepsis," and "acute kidney injury." The task is formulated as a series of independent binary classifications, one for each of the three problems. For each problem individually, the system is given only the S+O sections of a SOAP note and must determine if the patient has that specific condition ("Yes" or "No")

\subsection{The Multi-Agent System Architecture}
Our system consists of agents with defined roles and capabilities interacting in a centrally-coordinated workflow. The S+O section of a patient's SOAP note constitutes the shared "environment," the external context that all agents perceive and upon which they act \cite{tran2025multiagentcollaborationmechanismssurvey}. The architecture is composed of several key components.

\subsubsection{System Components}

\paragraph{Base Agent Component}
All agents in our system, including the Manager, the specialists, and the baseline zero-shot chain-of-thought agent, are built upon a common base component. This foundational component provides shared core functionalities essential for their operation, including LLM inference and a token-aware context management mechanism. To handle long clinical notes without exceeding the model's context window, this mechanism dynamically summarizes the conversation history. When the token count surpasses a threshold (70\% of the context limit), the longest message in the history is summarized using a low-temperature (t=0.1) LLM call to ensure factual preservation.

\paragraph{Manager Agent}
The Manager agent is the central coordinator of the entire diagnostic process. Its primary responsibilities are:
\begin{enumerate}
\item \textbf{Dynamic Role Creation:} In a two-step process, the Manager first queries the LLM to identify a set number of relevant medical specialties based on the note's content and the diagnostic question. It then prompts the LLM again to generate a concise list of key expertise areas for each identified specialty.
\item \textbf{Orchestration and Moderation:} The Manager instantiates the specialist agents, tasks them with the analysis, and facilitates the multi-round debate among them. It does not participate in the debate itself but monitors the process.
\item \textbf{Consensus and Final Decision:} After each round, it checks for consensus. If no consensus is reached within the specified rounds, it orchestrates a team re-assignment or acts as a final aggregator to make a decision.
\end{enumerate}

\paragraph{Dynamic Specialist Agents}
These specialist agents form the core analytical team of the system. Each agent is dynamically instantiated by the Manager and assigned a specific medical specialty (e.g., 'Cardiologist', 'Nephrologist') based on an analysis of both the clinical note's content and the specific diagnostic question (e.g., sepsis). Their expertise is activated through a detailed prompt instructing them to reason as if they were a human specialist. This design intentionally introduces a diversity of expert viewpoints to foster a more holistic analysis and enable collaborative error correction \cite{tran2025multiagentcollaborationmechanismssurvey, wang2024mixtureofagentsenhanceslargelanguage}.

\begin{figure}[h]
 \centering
 \includegraphics[width=\linewidth]{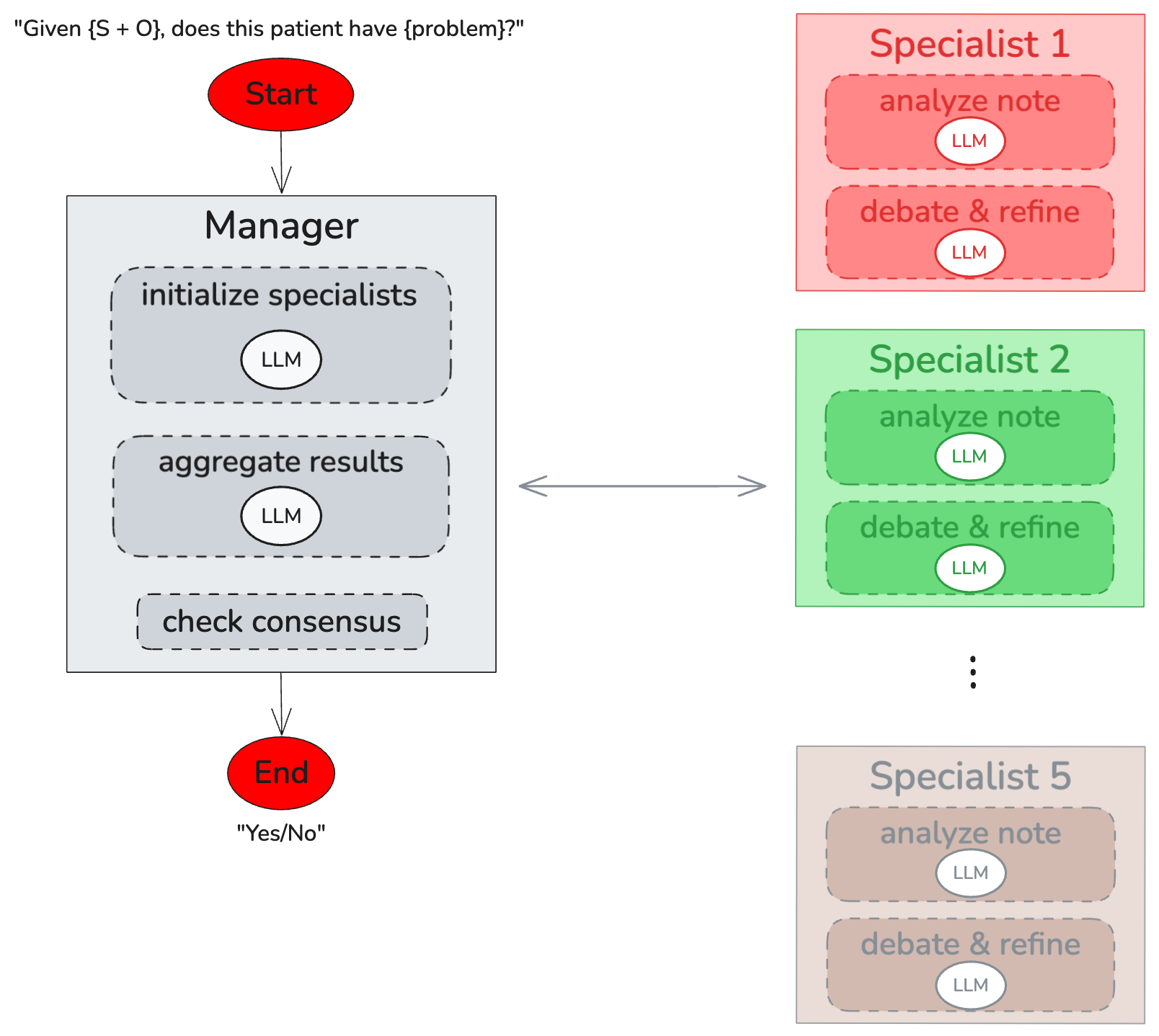}
  \caption{Components of our MAS. A Manager LLM (left) dynamically assembles a team of five Specialist agents (right), gathers their independent analyses, facilitates up to three rounds of iterative discussion, and allows for up to two team reassignments before reaching a final decision.}
  \label{fig:overview}
\end{figure}

Figure \ref{fig:overview} illustrates the components of our proposed multi-agent system, showing how the allowed behaviors of each agent work together to solve the diagnostic question.

\subsection{The Collaborative Reasoning Workflow}
The system employs a multi-layered, manager-coordinated protocol to reach a diagnostic conclusion. The workflow for a single clinical note proceeds through the following steps:

\paragraph{Step 1: Dynamic Team Assembly}
The Manager initiates the process by assembling a team of 5 specialist agents. It dynamically determines the most relevant specialties and their expertise as described above.

\paragraph{Step 2: Independent Analysis (Round 1)}
The Manager tasks each of the 5 specialists with independently analyzing the S+O text. To maximize efficiency, these concurrent analyses are executed in parallel. Each agent must provide its reasoning and a binary choice ("Yes" or "No") regarding the presence of the specified clinical problem.

\paragraph{Step 3: Iterative Debate and Consensus Check (Rounds 2-3)}
If no consensus is reached in Round 1, the debate phase begins. For up to 3 total rounds, the following occurs:
\begin{itemize}
\item Each agent is provided with the reasoning and conclusions from all other agents in the previous round.
\item Each agent is prompted to reconsider its own analysis in light of its peers' arguments, refine their conclusions, and submit a potentially revised choice.
\item After each round, the Manager checks if a consensus has been met. A consensus is defined as 80\% or more of the agents agreeing on the same choice. If consensus is reached, the process terminates, and that choice is recorded as the final answer.
\end{itemize}

\paragraph{Step 4: Team Re-Assignment (Hierarchical Debate)}
If the initial team of specialists fails to reach a consensus after 3 rounds, the Manager disbands the team. It then summarizes the previous team's final arguments and initiates the entire process again, starting from Step 1 with a brand new, dynamically generated team of specialists. This re-assignment can occur up to a maximum of 2 times.

\paragraph{Step 5: Managerial Aggregation (Fallback)}
If no consensus is reached after all team assignment attempts have been exhausted, the Manager enters a fallback mode. It reviews the entire history of debates from all teams and acts as a final, top-level aggregator to synthesize a single, definitive answer.

\subsection{Experimental Setup}

\paragraph{Baseline}
The baseline is a single, zero-shot chain-of-thought agent, prompted to perform the same task without any collaborative elements. As it is built from the same base component, it shares the same core inference settings as the multi-agent system.

\paragraph{MAS Configuration}
The primary configuration evaluated, as detailed in the workflow above, consists of a Manager orchestrating a team of 5 dynamically assigned specialist agents, with a consensus threshold of 0.8, a maximum of 3 debate rounds per team, and a maximum of 2 team re-assignment attempts.

\noindent
To probe the design space, we also tried three alternative team layouts
(Generic, Hybrid, Static–Dynamic); their full definitions and results
appear in Appendix~\ref{sec:ablations}.

\paragraph{Evaluation Metrics}
We evaluate our system's performance by treating the identification of each clinical problem as a binary classification task. For every SOAP note, the system's prediction ('Yes' or 'No') is compared against the ground-truth label derived from our dataset. This comparison yields one of four outcomes:

\begin{itemize}
    \item \textbf{True Positive (TP):} The system correctly predicts 'Yes' for a patient who has the clinical problem.
    \item \textbf{True Negative (TN):} The system correctly predicts 'No' for a patient who does not have the problem.
    \item \textbf{False Positive (FP):} The system incorrectly predicts 'Yes' for a patient who does not have the problem (a Type I error or "false alarm").
    \item \textbf{False Negative (FN):} The system incorrectly predicts 'No' for a patient who does have the problem (a Type II error or "miss").
\end{itemize}

From these counts, we calculate four key metrics to provide a comprehensive assessment of the model's clinical utility.

\begin{description}
    \item[Precision] (or Positive Predictive Value) measures the accuracy of positive predictions. It answers the question: "Of all the patients the system flagged, how many actually had the disease?" In a clinical setting, high precision is vital for minimizing unnecessary treatments, tests, and patient anxiety that result from false alarms. It is calculated as:
    $$ \text{Precision} = \frac{TP}{TP + FP} $$

    \item[Recall] (or Sensitivity) measures the model's ability to identify all actual positive cases. It answers the question: "Of all the patients who truly had the disease, how many did the system find?" High recall is critical in diagnostics, as failing to identify a condition (a false negative) can lead to delayed treatment and poor patient outcomes. It is calculated as:
    $$ \text{Recall (Sensitivity)} = \frac{TP}{TP + FN} $$

    \item[Specificity] measures the model's ability to correctly identify negative cases. It is complementary to recall and answers: "Of all the patients who were healthy, how many did the system correctly clear?" High specificity is important for preventing healthy individuals from undergoing unnecessary and costly follow-up procedures. It is calculated as:
    $$ \text{Specificity} = \frac{TN}{TN + FP} $$

    \item[F1-Score] is the harmonic mean of precision and recall, providing a single score that balances the two. It is particularly useful when dealing with imbalanced datasets, which are common in medicine where the number of healthy patients often far exceeds the number of sick patients. A high F1-score indicates that the model has a good balance between not making false alarms and not missing true cases. It is calculated as:
    $$ \text{F1-Score} = 2 \times \frac{\text{Precision} \times \text{Recall}}{\text{Precision} + \text{Recall}} $$
\end{description}

\paragraph{Implementation Details}
Experiments were conducted using the \texttt{meta-llama/Llama-3-70B-Instruct} model. The model was served locally from a server equipped with four NVIDIA A40 GPUs, using the vLLM~\cite{kwon2023efficientmemorymanagementlarge} library for efficient inference. We configured vLLM with a tensor-parallel size of 4 to distribute the model across all GPUs. The vLLM instance exposed an OpenAI-compatible API endpoint, allowing our system to interface with the local model using the Python AsyncOpenAI library. 

For all diagnostic reasoning tasks—including the baseline analysis, specialist role creation, and debate rounds—the LLM temperature was set to 0.5. The only exception was the context summarization module, which used a more deterministic temperature of 0.1. 

The entire pipeline was designed for parallel execution. Each clinical note was processed as an independent asynchronous task, and within each task, the analyses by individual specialist agents were executed concurrently using Python’s asyncio library. To ensure reliable parsing of all agent responses, structured JSON outputs from the LLM were enforced using the Pydantic library and the lm-format-enforcer backend\footnote{Available at \url{https://github.com/noamgat/lm-format-enforcer}}, integrated via vLLM.

\section{Results}
We evaluated the performance of our Dynamic Specialist MAS against the single-agent baseline on the task of identifying "congestive heart failure," "acute kidney injury," and "sepsis." The results, summarized in Table \ref{tab:main_results}, show that the Dynamic Specialist MAS generally outperformed the Baseline (zero-shot chain-of-thought) model. The overall F1-score, which balances precision and recall, improved from a macro-average of 0.493 for the baseline to 0.502 for our multi-agent system. 

\textbf{Extended results.}
Beyond the single–baseline comparison reported in Table~\ref{tab:main_results}, we evaluated three alternative team configurations—\emph{Generic}, \emph{Hybrid}, and \emph{Static–Dynamic}. Appendix~\ref{sec:ablations} shows that Dynamic Specialist MAS is consistently competitive and often the top performer, although specific runs or
disease categories occasionally favour another variant. This overall robustness, coupled with its fully data‑driven
specialist recruitment, is why we keep it as the focal configuration.

\begin{table}[h]
  \caption{Performance comparison of the Baseline (ZSCOT) and Dynamic Specialist MAS.}
  \label{tab:main_results}
  \centering
  \scriptsize
  \setlength{\tabcolsep}{3pt}  
  \begin{tabular*}{\linewidth}{@{\extracolsep{\fill}}lrrrrcccc}
    \toprule
    \multirow{2}{*}{Method} & \multicolumn{4}{c}{Counts} & \multicolumn{4}{c}{Metrics}\\
    \cmidrule(lr){2-5}\cmidrule(lr){6-9}
           & TP & TN & FP & FN & Precision & Recall & Specificity & F1\\
    \midrule
    \multicolumn{9}{l}{\textbf{Congestive Heart Failure}}\\
    Baseline ZSCOT            &  26 & 279 &  49 &  66 & 0.347 & 0.283 & 0.851 & 0.311\\
    Dynamic Specialist MAS &  26 & 286 &  42 &  66 & 0.382 & 0.283 & 0.872 & 0.325\\
    \midrule
    \multicolumn{9}{l}{\textbf{Acute Kidney Injury}}\\
    Baseline ZSCOT            &  80 & 200 &  86 &  54 & 0.482 & 0.597 & 0.699 & 0.533\\
    Dynamic Specialist MAS &  83 & 197 &  89 &  51 & 0.483 & 0.619 & 0.689 & 0.542\\
    \midrule
    \multicolumn{9}{l}{\textbf{Sepsis}}\\
    Baseline ZSCOT            &  96 & 213 &  78 &  33 & 0.552 & 0.744 & 0.732 & 0.634\\
    Dynamic Specialist MAS &  99 & 209 &  82 &  30 & 0.547 & 0.767 & 0.718 & 0.639\\
    \midrule
    \multicolumn{9}{l}{\textbf{Macro‑Average}}\\
    Baseline ZSCOT            & & & & & 0.460 & 0.541 & 0.761 & 0.493\\
    Dynamic Specialist MAS & & & & & 0.471 & 0.556 & 0.760 & 0.502\\
    \bottomrule
  \end{tabular*}
\end{table}

\begin{figure}[h]
 \centering
 \includegraphics[width=\linewidth]{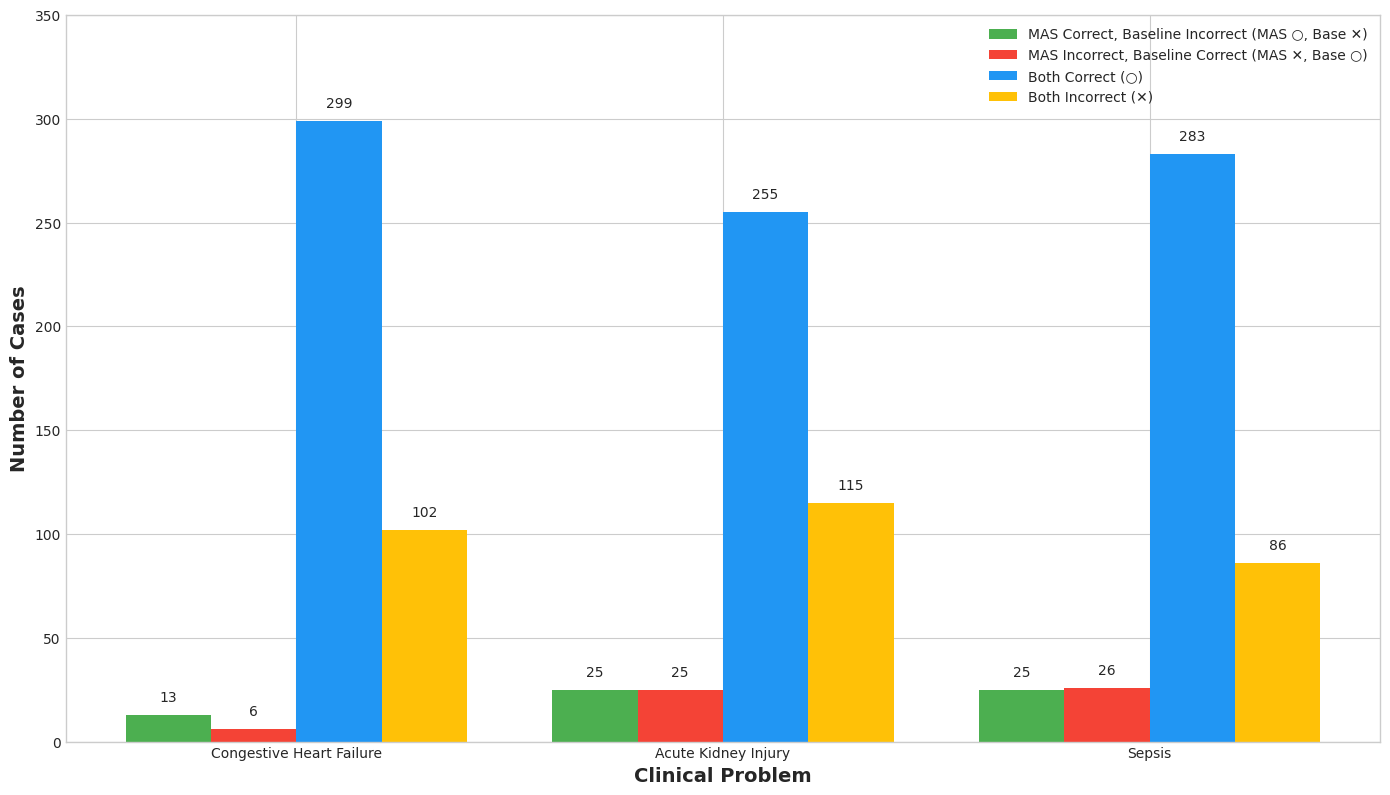}
  \caption{Distribution of note‑level prediction outcomes, grouped by clinical problem and by which methods were correct.} 
  \label{fig:bucket_plot}
\end{figure}

To clarify how these aggregate metrics map to individual clinical notes, Figure~\ref{fig:bucket_plot} breaks down each prediction according to which method was correct.

Table~\ref{tab:main_results} weights false positives and false negatives differently, whereas Figure~\ref{fig:bucket_plot} simply counts notes. 
For sepsis, MAS converts three of the baseline’s false negatives into true positives (TP: 96 → 99; FN: 33 → 30) while adding four false alarms (FP: 78 → 82). These changes lift the recall from 0.744 to 0.767. While this comes with a slight drop in precision (0.552 → 0.547), the F1-score still improves (0.634 → 0.639). This trade-off is highly desirable in a clinical context, where the cost of a false negative (missing a sick patient) is typically much higher than the cost of a false positive. It is particularly noteworthy that the largest gain in recall was achieved for sepsis, a condition where early recognition and treatment are critical for patient outcomes.

In Figure\ref{fig:bucket_plot}, however, one extra false positive (MAS X, Baseline O) raises the red bar exactly as much as one rescued positive case (MAS O, Baseline X) raises the green bar. Because the figure is \emph{case‑count neutral} while the metrics are \emph{error‑type aware}, the red and green bars can appear similar in height even when MAS delivers a better overall balance of precision and recall. Viewing the confusion-matrix counts alongside the plot resolves this apparent contradiction.

\subsection{Qualitative case analysis: when collaboration helps—and when it hurts}

To understand the cases where MAS and the Baseline disagree (green and red bars in Figure~\ref{fig:bucket_plot}), we randomly sampled several of these cases and manually reviewed their chat histories. Our analysis revealed a key duality in the collaborative process. The dynamic assignment of specialists allows for the integration of diverse perspectives, which often corrects the Baseline's initial, less thorough assessments by identifying subtle clinical clues. However, this same process can sometimes introduce flawed group dynamics, where one or two vocal agents can cause the panel to fixate on misleading evidence, leading to an incorrect consensus.

The following cases illustrate these two patterns in practice.

    \begin{itemize}
        \item \textit{Example of 'MAS O, Baseline X'}
        \begin{itemize}
            \item \textbf{HADM ID:} 191928
            \item \textbf{Problem:} "Acute Kidney Injury"
            \item \textbf{Gold / MAS / Baseline:} "No" / "No" / "Yes"
            \item \textbf{Decisive clinical facts in the note:} Creatinine 1.3 → 1.4 mg/dL; Urine 430 mL/24 h
            \item \textbf{Reasoning of Baseline:} "The patient's creatinine level is 1.4 mg/dl, which is elevated. [...] Given the elevated creatinine level and the clinical context, it is reasonable to suspect that the patient may have acute kidney injury, [...]"
            \item \textbf{Reasoning of Internal Medicine Specialist in MAS, at round 1 (voted "No"):} "the increase from 1.3 to 1.4 mg/dl is less than 0.3 mg/dl and does not meet the 50\% increase criterion." 
            \item \textbf{Reasoning of Nephrologist in MAS, at round 1 (voted "Yes"):} "[...] Given the elevated creatinine level and decreased urine output, these findings suggest impaired kidney function consistent with acute kidney injury."
            \item \textbf{Reasoning of Nephrologist in MAS, at round 2 (voted "No"):} "Upon reviewing the reasoning provided by my peers, [...] the increase in creatinine from 1.3 to 1.4 mg/dl, although indicative of some kidney dysfunction, does not meet the specific criteria for AKI." 
        \end{itemize}
        \vspace{0.5em}
        \begin{itemize}
            \item \textbf{HADM ID:} 105852
            \item \textbf{Problem:} "Congestive Heart Failure"
            \item \textbf{Gold / MAS / Baseline:} "No" / "No" / "Yes"
            \item \textbf{Decisive clinical facts in the note:} Fluid Balance: -181 mL, -480 mL; Respiratory Rate (RR): 15/min (unlabored); Edema: Trace
            \item \textbf{Reasoning of Baseline:} "[...] The patient's medication list includes furosemide (Lasix), a diuretic commonly used to treat fluid overload in CHF, suggesting that the patient may have a condition requiring diuresis, such as CHF or significant fluid retention. [...]"
            \item \textbf{Reasoning of Cardiologist in MAS, at round 1 (voted "Yes"):} "[...] The patient's physical examination reveals a neck with a jugular venous pressure (JVP) of 10 cm, which is elevated and suggests fluid overload. The lungs show diminished breath sounds and crackles at the bases, indicating pulmonary congestion, another sign of CHF. [...]"
            \item \textbf{Reasoning of Nephrologist in MAS, at round 1 (voted "No"):} "[...] However, the fluid balance data shows a negative balance (-181 ml and -480 ml), suggesting that the patient is not in significant fluid overload. Overall, while there are some signs that could be consistent with congestive heart failure, such as pulmonary congestion and slightly elevated JVP, they are not conclusive, and other indicators like significant edema and positive fluid balance are not present."
            \item \textbf{Reasoning of Pulmonologist in MAS, at round 1 (voted "No"):} "[...] but the absence of significant respiratory distress, orthopnea, or paroxysmal nocturnal dyspnea in the note suggests that congestive heart failure (CHF) may not be the primary issue. [...]"
            \item \textbf{Reasoning of Cardiologist in MAS, at round 2 (voted "No"):} "Considering the input from all specialties, [...] The absence of key indicators such as significant peripheral edema, more pronounced respiratory distress, and higher JVP suggests that CHF might not be the primary issue at hand. Therefore, revising my initial assessment, I conclude that the patient does not have congestive heart failure based on the provided information."
        \end{itemize}
        \vspace{0.5em}
        \begin{itemize}
            \item \textbf{HADM ID:} 103771
            \item \textbf{Problem:} "Sepsis"
            \item \textbf{Gold / MAS / Baseline:} "No" / "No" / "Yes"
            \item \textbf{Decisive clinical facts in the note:} WBC 10.6 k/ul (within normal limits); Tmax 37.1°C (no fever); "blood, urine, and sputum cultures pending with no growth to date."
            \item \textbf{Reasoning of Baseline:} "[...]The patient is also on broad-spectrum antibiotics (meropenem and ciprofloxacin), which suggests a suspected infection. Additionally, the patient has a high white blood cell count (10.6 k/ul) and a slightly elevated temperature (37.1\u00b0C), which are signs of a potential infection. [...] the patient's clinical presentation is consistent with a potential infection and organ dysfunction, which are key components of sepsis."
            \item \textbf{Reasoning of Pulmonologist in MAS, at round 1 (voted "Yes"):} "The patient's note shows a high white blood cell count (WBC) of 10.6 k/ul, which may indicate an infection. [...] the combination of a suspected infection, fever, and respiratory symptoms suggests that the patient may have sepsis."
            \item \textbf{Reasoning of Infectious Disease Specialist in MAS, at round 1 (voted "No"):} "[...] However, the patient's white blood cell count (WBC) is 10.6 k/ul, which is within the normal range, and the patient is not showing signs of fever (temperature is 36.4°C or 97.6°F). The patient's blood, urine, and sputum cultures are pending with no growth to date. While the patient is showing some signs of a potential infection, the lack of fever, normal WBC count, and pending culture results make it difficult to definitively diagnose sepsis at this time."
            \item \textbf{Reasoning of Pulmonologist in MAS, at round 2 (voted "No"):} "[...]I notice that the Infectious Disease Specialist and Critical Care Medicine Specialist have pointed out that the patient's white blood cell count (WBC) is within the normal range (10.6 k/ul) [...]. The Nephrologist has also noted that the Systemic Inflammatory Response Syndrome (SIRS) criteria are not fully met. [...] the lack of confirmed hypotension, positive cultures, and elevated WBC count makes it difficult to definitively diagnose sepsis. The Cardiologist's opinion that the patient has sepsis is based on the presence of a suspected infection, fever, and respiratory symptoms, but this diagnosis is not universally agreed upon by the other specialists. [...]"
        \end{itemize}
        \vspace{0.5em}
        \item \textit{Example of 'MAS X, Baseline O'}
        \begin{itemize}
            \item \textbf{HADM ID:} 189576
            \item \textbf{Problem:} "Acute Kidney Injury"
            \item \textbf{Gold / MAS / Baseline:} "No" / "Yes" / "No"
            \item \textbf{Decisive clinical facts in the note:} Cr 1.3 → 1.0 mg/dL (down); UOP 310 mL/24 h; weight unknown
            \item \textbf{Reasoning of Baseline:} "the serum creatinine levels [...] do not show a significant increase that would typically indicate AKI." 
            \item \textbf{Reasoning of Cardiologist in MAS, at round 1 (voted "No"):} "[...] the creatinine levels are 1.3, 1.2, and 1.0 mg/dl, which, although slightly elevated, do not show a significant increase (> 0.3 mg/dL within 48 hours or 1.5 times baseline within 7 days) that is typically used to diagnose AKI. [...] we cannot definitively say the patient has AKI without more specific data on the baseline creatinine or a more significant increase in creatinine levels over time."
            \item \textbf{Reasoning of Cardiologist in MAS, at round 2 (voted "Yes"):} "Upon reviewing the reasoning provided by my peers from different specialties, several key points emerge that warrant reconsideration of my initial assessment. The Nephrologist, Critical Care Medicine Specialist, and Urologist all highlight the importance of the patient's marginal urine output and elevated serum creatinine levels as indicators of potential acute kidney injury (AKI). [...] the preponderance of evidence from the patient's marginal urine output, elevated creatinine levels, and clinical context [...] I am revising my initial response to align with the majority opinion that the patient does have indicators suggestive of acute kidney injury." 
        \end{itemize}
        \vspace{0.5em}
        \begin{itemize}
            \item \textbf{HADM ID:} 184837
            \item \textbf{Problem:} "Congestive Heart Failure"
            \item \textbf{Gold / MAS / Baseline:} "No" / "Yes" / "No"
            \item \textbf{Decisive clinical facts in the note:} Jugular Venous Pressure (JVP): Not elevated; Edema: None; Fluid Balance: -405 mL, -1,110 mL
            \item \textbf{Reasoning of Baseline:} "Although the patient has been weaned off a nitro drip, which is sometimes used to treat CHF, and has bibasilar rales, which can be a sign of fluid overload, the patient's jugular venous pressure (JVP) is not elevated, and the cardiac exam is otherwise normal (RRR, normal S1, S2, no murmurs). [...] Overall, the patient's presentation is not strongly suggestive of congestive heart failure."
            \item \textbf{Reasoning of Nephrologist in MAS, at round 1 (voted "No"):} "[...]the absence of elevated jugular venous pressure (JVP not elevated), and the description of the lungs as having 'normal respiratory effort' with only 'bibasilar rales' suggest against severe pulmonary edema typically seen in CHF. Furthermore, the cardiac examination findings of 'rrr, normal s1, s2. no m/r/g' (regular rate and rhythm, normal first and second heart sounds, no murmurs, rubs, or gallops) do not strongly support CHF. [...]"
            \item \textbf{Reasoning of Cardiologist in MAS, at round 1 (voted "Yes"):} "[...]The presence of bibasilar rales in the lungs, indicating fluid accumulation, is a sign often associated with congestive heart failure. [...] including the use of diuretics, signs of fluid overload (bibasilar rales), and elevated troponin-T, these factors lean towards the diagnosis of congestive heart failure."
            \item \textbf{Reasoning of Nephrologist in MAS, at round 2 (voted "Yes"):} "[...] The cardiologist, [...] citing the use of furosemide (Lasix), the presence of bibasilar rales indicating fluid accumulation in the lungs, and the elevated troponin-T level suggesting myocardial injury. [...] As a nephrologist, my primary focus is on renal function and fluid balance, but acknowledging the interdisciplinary insights, I recognize that the patient's condition may indeed be consistent with CHF, especially considering the cardiac and pulmonary findings."
        \end{itemize}
        \vspace{0.5em}
        \begin{itemize}
            \item \textbf{HADM ID:} 185625
            \item \textbf{Problem:} "Sepsis"
            \item \textbf{Gold / MAS / Baseline:} "No" / "Yes" / "No"
            \item \textbf{Decisive clinical facts in the note:} Tmax 37.6°C (no significant fever); No clear source of infection identified
            \item \textbf{Reasoning of Baseline:} "[...]However, the patient's temperature is not significantly elevated [...] The patient's hypoxia, tachycardia, and pulmonary edema could be related to various conditions, including cardiac issues or other non-infectious causes. [...] Without more definitive evidence of an infection or a clearer indication that the patient's condition is due to sepsis rather than another cause, it's challenging to conclusively diagnose sepsis based on the provided information."
            \item \textbf{Reasoning of Pulmonologist in MAS, at round a (voted "No"):} "[...]there is no clear source of infection mentioned in the note. [...] While the patient's condition is concerning and requires close monitoring, the available information does not definitively confirm sepsis."
            \item \textbf{Reasoning of Critical Care Medicine Specialist in MAS, at round 1 (voted "Yes"):} "Sepsis is typically defined by the presence of a suspected or documented infection, plus signs of a systemic inflammatory response syndrome (SIRS), [...] The patient has a heart rate of 109 bpm, which is elevated, and a white blood cell count of 13.8 k/ul, which is also elevated. [...]"
            \item \textbf{Reasoning of Pulmonologist in MAS, at round 2 (voted "Yes"):} "[...] They highlight the elevated white blood cell count, suspected infection (as evidenced by antibiotic use), and signs of organ dysfunction (such as worsening pulmonary edema and desaturation episodes) as indicators of sepsis. The Critical Care Medicine Specialist also notes that the patient's heart rate and white blood cell count meet the criteria for Systemic Inflammatory Response Syndrome (SIRS), which is a key component of sepsis.[...]Therefore, taking into account the input from my peers and re-evaluating the patient's note, I am revising my initial answer to reflect the possibility of sepsis."
        \end{itemize}
    \end{itemize}

These cases highlight two key patterns. First, in successful collaborations (MAS O, Baseline X), specialists correct initial errors by bringing in their unique expertise. For example, an Internal Medicine specialist might focus on strict diagnostic criteria for AKI, while a Nephrologist and Pulmonologist might provide a broader context for Congestive Heart Failure, preventing a diagnosis based on a single piece of evidence like medication use. This collaboration ensures a more thorough and accurate assessment.

Second, in failed collaborations (MAS X, Baseline O), a form of "groupthink" can occur. Specialists sometimes change their correct initial assessment to align with a majority opinion, even when that opinion is based on less critical evidence. This was seen when agents focused on marginal urine output for AKI despite a decreasing creatinine trend, or on secondary signs for CHF and Sepsis while overlooking the absence of core indicators like elevated JVP or significant fever. This shows that while collaboration can add valuable perspectives, it also risks amplifying biases if not carefully managed.

\subsection{Analysis of Agent Collaboration Dynamics}
To understand the internal mechanics of the MAS, we analyzed the debate transcripts to quantify how agents interacted. We focused on three key aspects: the complexity of the debate, the composition of the specialist teams, and the influence of individual specialist roles.

\paragraph{Debate Complexity and Efficiency}
Our analysis shows that despite being equipped for complex, multi-round debates and team re-assignments, the system resolved most cases with remarkable efficiency. As shown in Table \ref{tab:debate_stats}, the average number of debate rounds required to reach a final decision was very low, ranging from 1.06 for Congestive Heart Failure to 1.21 for Sepsis when the final answer was incorrect. Crucially, the need to re-assign a new specialist team was exceedingly rare, with the average number of panels per case hovering just above 1.0. This indicates that for the vast majority of notes, the initial, dynamically-selected team of five specialists was sufficient to reach a stable consensus quickly, often after just one round of exchanging views. The system's more complex hierarchical features, while important for handling difficult edge cases, were not a common requirement. It is worth noting that our current implementation is homogeneous, with all agents powered by the same underlying LLM. A promising direction for future research would be to explore a heterogeneous agent team, where different specialists are backed by different LLM models, potentially introducing another layer of cognitive diversity.

\begin{table}[h]
\centering
\caption{Average debate rounds and panel re-assignments required for the MAS to reach a final decision, grouped by correctness of the outcome.}
\label{tab:debate_stats}
\scriptsize 
\begin{tabular}{lcccc}
\toprule
\multirow{2}{*}{Problem} & \multicolumn{2}{c}{Avg. Debate Rounds} & \multicolumn{2}{c}{Avg. Panels Assigned} \\
\cmidrule(lr){2-3} \cmidrule(lr){4-5}
& Correct & Incorrect & Correct & Incorrect \\
\midrule
Congestive Heart Failure & 1.06 & 1.06 & 1.01 & 1.01 \\
Acute Kidney Injury   & 1.13 & 1.14 & 1.04 & 1.05 \\
Sepsis                & 1.15 & 1.21 & 1.03 & 1.04 \\
\bottomrule
\end{tabular}
\end{table}

\begin{figure}[h!]
\centering
\includegraphics[width=\linewidth]{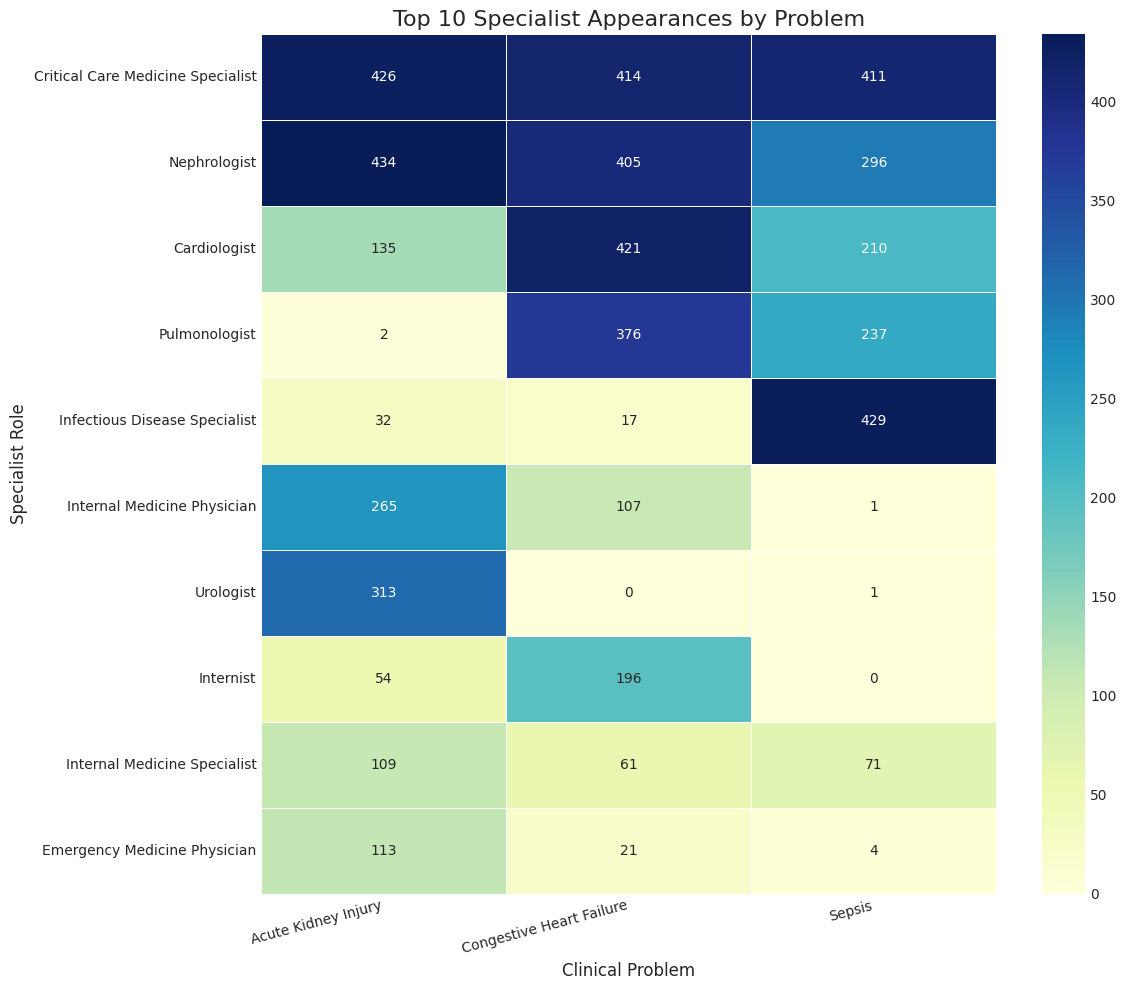}
\caption{Heatmap of the top 10 most frequent specialist agent appearances across the three clinical problems. Color intensity corresponds to the number of times a specialist was recruited for a given problem.}
\label{fig:specialist_heatmap_top10}
\end{figure}

\paragraph{Specialist Participation and Problem Complexity}
The system's dynamic role creation resulted in specialist teams tailored to each clinical problem, reflecting real-world clinical logic. For \textbf{Acute Kidney Injury}, the most frequently recruited agent was the \textbf{Nephrologist}. For \textbf{Congestive Heart Failure}, it was the \textbf{Cardiologist}, and for \textbf{Sepsis}, the \textbf{Infectious Disease Specialist}. Interestingly, the diversity of specialists recruited appears to correlate with the clinical complexity of the condition. \textbf{Sepsis}, a systemic condition affecting multiple organ systems, required the most diverse panel, with 47 unique specialist roles appearing in the debates. This was followed by \textbf{Acute Kidney Injury} (43 unique roles) and \textbf{Congestive Heart Failure} (37 unique roles). This suggests the Manager agent effectively identified the broader range of expertise needed for more complex, systemic problems. A detailed visualization of the top 10 specialist roles is presented in Figure~\ref{fig:specialist_heatmap_top10}. A future configuration could involve a hybrid approach where core specialists are statically included based on the target problem, while the rest of the panel is still assigned dynamically to adapt to the specific details of the note.

\paragraph{Identifying Pivotal Specialist Roles}
To quantify the influence of each specialist, we defined a "decisiveness score." An agent was considered to have made a decisive pivot if it changed its vote during the debate, and its new vote matched the final consensus of the system. The score is the ratio of these pivotal moments to the total number of times the role appeared, calculated as:

\begin{equation} \label{eq:decisiveness}
\text{Decisiveness} = \frac{\text{Number of Pivotal Changes}}{\text{Total Appearances of the Role}}
\end{equation}

We focused only on roles with a substantial number of appearances (n > 100), as a rarely-recruited specialist could achieve a misleadingly high score from a single pivotal case. Table \ref{tab:pivotal_roles} summarizes the most influential of these frequently appearing roles. The results show that roles most closely aligned with the target disease consistently acted as key influencers. For instance, in AKI cases, the \textbf{Emergency Medicine Physician} had a high decisiveness score of 0.150, reflecting their critical role in swaying the team.

\begin{table}[h]
  \caption{Most influential specialist roles based on decisiveness score. The score measures how often a role's change in vote aligned with the final system outcome. $p$ = pivot count, $n$ = total appearances.}
  \label{tab:pivotal_roles}
  \centering
  \small
  \begin{tabular}{llccc}
    \toprule
    Problem & Role & $p$ & $n$ & Decisiveness\\
    \midrule
    \multirow{5}{*}{CHF}
      & Internist                         & 12 & 196 & 0.061\\
      & Cardiologist                      & 19 & 421 & 0.045\\
      & Nephrologist                      & 15 & 405 & 0.037\\
      & Pulmonologist                     & 12 & 376 & 0.032\\
      & Critical Care Med.\ Spec.         &  9 & 414 & 0.022\\
    \midrule
    \multirow{5}{*}{AKI}
      & Emergency Med.\ Physician         & 17 & 113 & 0.150\\
      & Intensive Care Med.\ Spec.        & 12 & 101 & 0.119\\
      & Cardiologist                      & 13 & 135 & 0.096\\
      & Urologist                         & 30 & 313 & 0.096\\
      & Critical Care Med.\ Spec.         & 37 & 426 & 0.087\\
    \midrule
    \multirow{5}{*}{Sepsis}
      & Pulmonologist                     & 25 & 237 & 0.105\\
      & Cardiologist                      & 22 & 210 & 0.105\\
      & Critical Care Med.\ Spec.         & 36 & 411 & 0.088\\
      & Infectious Disease Specialist     & 36 & 429 & 0.084\\
      & Nephrologist                      & 22 & 296 & 0.074\\
    \bottomrule
  \end{tabular}
\end{table}

\section{Discussion}
Our results confirm our primary hypothesis: a collaborative multi-agent architecture with dynamically assigned specialists outperforms a single-LLM baseline in identifying clinical problems from S+O notes. The quantitative improvements, particularly the increased recall, highlight the clinical relevance of our approach. This suggests that the MAS is better at avoiding high-cost false negatives, a crucial feature for any diagnostic aid.

However, the most significant insights come from analyzing how these improvements were achieved. The qualitative case analysis reveals that the debate mechanism frequently functions as an effective error-correction process. In successful cases, specialists introduced domain-specific knowledge (e.g., precise diagnostic criteria for AKI) or a more holistic perspective (e.g., weighing negative fluid balance against other signs in CHF), correcting a simplistic or narrowly-focused initial assessment from the baseline model. This process of "inter-agent feedback" \cite{talebirad2023multiagentcollaborationharnessingpower} allows the system to surface and weigh conflicting evidence before reaching a conclusion, aligning with findings from general MAS research that collective intelligence enhances response quality \cite{wang2024mixtureofagentsenhanceslargelanguage}.

Conversely, our analysis also uncovered the primary failure mode of the system: a form of "groupthink." In some cases, agents with an initially correct assessment changed their vote to align with a flawed majority, often swayed by less critical evidence. This reveals a key duality: the same collaborative mechanism that corrects individual errors can, under certain conditions, amplify them. This finding underscores the importance of the debate protocol's design and suggests that future work should explore mechanisms to mitigate this effect.

The analysis of agent collaboration dynamics further validates our design. The system's efficiency—resolving most cases in the first round without needing the hierarchical fallback of team re-assignment—demonstrates its practicality. The dynamic recruitment of specialists (e.g., Cardiologists for CHF, Nephrologists for AKI) and the correlation between specialist diversity and disease complexity (with sepsis requiring the widest range of experts) show that the Manager agent effectively tailors the team to the task. Furthermore, our "decisiveness score" provides an empirical method for identifying which roles are most influential, offering a path toward more sophisticated future architectures, such as weighted voting schemes.

\subsection{Limitations of the Study}
This study has several limitations. First, our dataset, while carefully curated for the reasoning task, is derived from a single database (MIMIC-III) and focuses on a specific type of clinical note. Our evaluation was limited to three clinical problems; the system's performance on the remaining 11 less frequent or more diagnostically nuanced problems is yet to be determined. 

Second, the system's capabilities are ultimately bounded by the foundation LLM, which has the potential for hallucination. While the consensus mechanism and iterative debate are designed to mitigate this, the "groupthink" phenomenon observed in our qualitative analysis indicates that systemic errors are still possible.

Third, MAS architectures are inherently more complex and computationally expensive than single-agent setups. Our study did not perform an exhaustive ablation of all parameters, such as the optimal number of agents or debate rounds, due to computational constraints. 

Finally, the outputs have not been validated by clinicians. While the debate transcripts appear to offer meaningful explanations, a crucial next step is to have medical experts assess their clinical relevance and the soundness of the agents' reasoning. This work should be viewed as a research prototype, not a clinical diagnostic tool.

\subsection{Future Work}
There are several exciting directions for future work. A significant next step is to mitigate the observed "groupthink" by designing more sophisticated debate protocols, perhaps by introducing an agent specifically designed to challenge the emerging consensus, or by giving more weight to initial, independent analyses. Another direction is to equip agents with "plugins" \cite{talebirad2023multiagentcollaborationharnessingpower} for tool use, such as the ability to query a medical knowledge base or a drug database via an API to perform real-time information retrieval (RAG). 

Building on our analysis of specialist roles, a hybrid team-building approach could be explored, where core specialists are statically included while others are assigned dynamically. We also plan to explore heterogeneous agent teams, where different specialists are powered by different underlying LLMs to increase cognitive diversity. Expanding the system to analyze a sequence of notes for a single patient would allow for tracking disease progression, moving closer to the comprehensive modeling seen in Agent Hospital \cite{li2025agenthospitalsimulacrumhospital}. Finally, integrating a human-in-the-loop for expert oversight during the debate and applying the framework to other clinical NLP tasks are important next steps.

\section{Conclusion}
In this work, we presented a collaborative multi-agent system designed to automate the identification of clinical problems from the Subjective and Objective sections of SOAP notes. By structuring the interaction between LLM agents to mimic a clinical consultation—featuring dynamic specialist assignment and a hierarchical debate protocol—we achieved more clinically relevant results than a standard single-agent approach. Our quantitative results showed improved performance, and our qualitative analysis provided crucial insights into the mechanics of this success, demonstrating how collaborative debate corrects errors while also revealing its potential pitfalls, such as groupthink. Furthermore, our analysis of the agent dynamics validated our design choices and offered empirical evidence of emergent, specialized behaviors within the system. This work bridges the gap between general MAS frameworks and practical healthcare applications. The modular, open-source architecture, which leverages a locally-hosted model to protect data privacy, serves as a blueprint for future research into more sophisticated, reliable, and transparent clinical AI assistants that could one day help clinicians in complex reasoning tasks.

\begin{acks}
This work was supported in part by the National Science Foundation under the Grants IIS-1741306 and IIS-2235548, and by the Department of Defense under the Grant DoD W91XWH-05-1-023.  This material is based upon work supported by (while serving at) the National Science Foundation.  Any opinions, findings, conclusions, or recommendations expressed in this material are those of the author(s) and do not necessarily reflect the views of the National Science Foundation.
\end{acks}

\bibliographystyle{ACM-Reference-Format}
\bibliography{main} 

\clearpage
\appendix

\section{Additional Experiments}
\label{sec:ablations}

This appendix supplements Section 4 by reporting ablations that motivated the final Dynamic Specialist MAS design while also suggesting avenues for deeper comparative work in future studies. Our multi-agent system is intentionally designed with a flexible and modular architecture to facilitate a broad exploration of different collaborative strategies. This design allows for easy modification of key parameters, enabling experiments with various configurations beyond the primary Dynamic Specialist MAS presented in the main paper. 

For instance, the framework can support teams composed of generic agents, specialists with pre-defined (static) roles based on the clinical problem, or mixed teams. Furthermore, the number of specialists can be a fixed integer or set to 'auto', allowing the Manager agent to determine the team size dynamically. This inherent customizability provides a robust foundation for future research into optimal multi-agent configurations for clinical NLP.

To that end, we explored several other architectural configurations to better understand the design space and tested the robustness of our findings against the inherent stochasticity of the LLM (temperature = 0.5) by conducting a second, identical experimental run.

The exploratory configurations tested include:
\begin{itemize}
\item \textbf{Generic MAS:} A team of 5 generic agents without any specified roles, representing a less specialized collaborative approach.

\item \textbf{Static-Dynamic MAS:} A team of 5 specialists where some roles were pre-defined based on the clinical problem and the others were dynamically assigned. The pre-defined (static) roles were:
\begin{itemize}
    \item For \textit{Congestive Heart Failure}: Cardiologist, Cardiac Electrophysiologist.
    \item For \textit{Acute Kidney Injury}: Nephrologist, Intensive Care Specialist.
    \item For \textit{Sepsis}: Infectious Disease Specialist, Intensive Care Specialist.
\end{itemize}
These static roles were determined by prompting a large language model (GPT-4.5) to list the most relevant medical specialties for each clinical problem; the top two recommendations were then selected.

\item \textbf{Hybrid MAS:} A mixed team of 5 agents, comprising 2 generic agents, 2 pre-defined (static) specialists, and 1 dynamically assigned specialist. This configuration used the same static specialist roles listed above.
\end{itemize}

The full results for all configurations across two independent runs are presented in Table \ref{tab:full_results_run1} and Table \ref{tab:full_results_run2}.

\begin{table*}
  \caption{Full Experimental Results for All Configurations (Run 1)}
  \label{tab:full_results_run1}
  \begin{tabular}{llcccc}
    \toprule
    Problem & Method & Precision & Recall & Specificity & F1-Score\\
    \midrule
    Congestive Heart Failure & Baseline ZSCOT & 0.347 & 0.283 & 0.851 & 0.311\\
                             & \textbf{Dynamic Specialist MAS} & 0.382 & 0.283 & 0.872 & 0.325\\
                             & Generic MAS & 0.354 & 0.304 & 0.845 & 0.327\\
                             & Hybrid MAS & 0.373 & 0.272 & 0.872 & 0.314\\
                             & Static-Dynamic MAS & 0.369 & 0.261 & 0.875 & 0.306\\
    \midrule
    Acute Kidney Injury & Baseline ZSCOT & 0.482 & 0.597 & 0.699 & 0.533\\
                        & \textbf{Dynamic Specialist MAS} & 0.483 & 0.619 & 0.689 & 0.542\\
                        & Generic MAS & 0.468 & 0.545 & 0.710 & 0.503\\
                        & Hybrid MAS & 0.469 & 0.612 & 0.675 & 0.531\\
                        & Static-Dynamic MAS & 0.468 & 0.604 & 0.678 & 0.528\\
    \midrule
    Sepsis & Baseline ZSCOT & 0.552 & 0.744 & 0.732 & 0.634\\
           & \textbf{Dynamic Specialist MAS} & 0.547 & 0.767 & 0.718 & 0.639\\
           & Generic MAS & 0.528 & 0.721 & 0.715 & 0.610\\
           & Hybrid MAS & 0.531 & 0.736 & 0.711 & 0.617\\
           & Static-Dynamic MAS & 0.511 & 0.721 & 0.694 & 0.598\\
    \midrule
    Macro-Average & Baseline ZSCOT & 0.460 & 0.541 & 0.761 & 0.493\\
                  & \textbf{Dynamic Specialist MAS} & 0.471 & 0.556 & 0.760 & 0.502\\
                  & Generic MAS & 0.450 & 0.523 & 0.757 & 0.480\\
                  & Hybrid MAS & 0.458 & 0.540 & 0.753 & 0.487\\
                  & Static-Dynamic MAS & 0.449 & 0.529 & 0.749 & 0.477\\
  \bottomrule
\end{tabular}
\end{table*}

\begin{table*}
  \caption{Full Experimental Results for All Configurations (Run 2)}
  \label{tab:full_results_run2}
  \begin{tabular}{llcccc}
    \toprule
    Problem & Method & Precision & Recall & Specificity & F1-Score\\
    \midrule
    Congestive Heart Failure & Baseline ZSCOT & 0.333 & 0.293 & 0.835 & 0.312\\
                             & \textbf{Dynamic Specialist MAS} & 0.381 & 0.261 & 0.881 & 0.310\\
                             & Generic MAS & 0.360 & 0.293 & 0.854 & 0.323\\
                             & Hybrid MAS & 0.388 & 0.283 & 0.875 & 0.327\\
                             & Static-Dynamic MAS & 0.385 & 0.272 & 0.878 & 0.318\\
    \midrule
    Acute Kidney Injury & Baseline ZSCOT & 0.474 & 0.545 & 0.717 & 0.507\\
                        & \textbf{Dynamic Specialist MAS} & 0.461 & 0.575 & 0.685 & 0.512\\
                        & Generic MAS & 0.465 & 0.537 & 0.710 & 0.498\\
                        & Hybrid MAS & 0.463 & 0.560 & 0.696 & 0.507\\
                        & Static-Dynamic MAS & 0.453 & 0.582 & 0.671 & 0.510\\
    \midrule
    Sepsis & Baseline ZSCOT & 0.517 & 0.705 & 0.708 & 0.597\\
           & \textbf{Dynamic Specialist MAS} & 0.542 & 0.752 & 0.718 & 0.630\\
           & Generic MAS & 0.551 & 0.760 & 0.725 & 0.638\\
           & Hybrid MAS & 0.533 & 0.752 & 0.708 & 0.624\\
           & Static-Dynamic MAS & 0.534 & 0.736 & 0.715 & 0.619\\
    \midrule
    Macro-Average & Baseline ZSCOT & 0.441 & 0.514 & 0.753 & 0.472\\
                  & \textbf{Dynamic Specialist MAS} & 0.461 & 0.529 & 0.761 & 0.484\\
                  & Generic MAS & 0.459 & 0.530 & 0.763 & 0.486\\
                  & Hybrid MAS & 0.461 & 0.532 & 0.760 & 0.486\\
                  & Static-Dynamic MAS & 0.457 & 0.530 & 0.755 & 0.482\\
  \bottomrule
\end{tabular}
\end{table*}

\paragraph{Performance of Exploratory Configurations}
The results across both runs justify our focus on the Dynamic Specialist MAS in the main paper.
No single MAS variant dominates every metric, but the
Dynamic Specialist MAS delivers the steadiest gains over the baseline across both runs.
The other multi‑agent configurations (Generic, Hybrid, and Static‑Dynamic) showed inconsistent performance.
For example, the Generic MAS's F1‑score was competitive for Sepsis in Run 2 but significantly under‑performed the baseline for Acute Kidney Injury in both runs.
This lack of reliable improvement suggests that while these architectures are possible within our framework, they may be less stable or require further tuning to be effective.

Beyond the numbers, Dynamic Specialist MAS also wins on convenience.
Because the model automatically picks its specialists for every note, we do not need to retune anything when we move to a new clinical problem. Teams with fixed, pre‑set roles cannot do that.

\paragraph{Replicability of Main Findings} Comparing the results for the Baseline and Dynamic Specialist MAS across both runs confirms the robustness of our central claim. We can assess this by looking at the macro-average F1-score (averaged across the three problems):
\begin{itemize}
\item In Run 1, the Dynamic Specialist MAS improved the macro-average F1-score from 0.493 (Baseline) to 0.502.
\item In Run 2, the Dynamic Specialist MAS improved the macro-average F1-score from 0.472 (Baseline) to 0.484.
\end{itemize}
In both independent runs, the dynamic multi-agent approach demonstrated a superior overall performance compared to the single-agent baseline. While there are minor fluctuations in scores for individual diseases (e.g., Congestive Heart Failure in Run 2), the consistent improvement in the macro-average F1-score indicates that the benefit of the collaborative architecture is replicable.

\end{document}